\documentclass[10pt,twocolumn,letterpaper]{article}

\usepackage{cvpr}              
\usepackage[dvipsnames]{xcolor}

\definecolor{cvprblue}{rgb}{0.21,0.49,0.74}
\usepackage[pagebackref,breaklinks,colorlinks,citecolor=cvprblue]{hyperref}
\usepackage{pifont}
\usepackage{colortbl}
\usepackage{url}
\usepackage{multirow}
\usepackage[accsupp]{axessibility}
\definecolor{First}{rgb}{0.95, 0.62, 0.61}
\definecolor{Second}{rgb}{0.97,0.81,0.63}
\definecolor{Third}{rgb}{1.0, 0.97, 0.70}
\def\paperID{1254} 
\def\confName{CVPR}
\def\confYear{2024}

\title{MMVP: A \textbf{M}ultimodal \textbf{M}oCap Dataset with \textbf{V}ision and \textbf{P}ressure Sensors}

\author{
He Zhang$^{*1}$, Shenghao Ren$^{*3}$, Haolei Yuan$^{*1}$,\\ Jianhui Zhao$^{1}$, Fan Li$^{1}$, Shuangpeng Sun$^{2}$, Zhenghao Liang$^{4}$,\\ Tao Yu$^{\dag2}$, Qiu Shen$^{\dag3}$, Xun Cao$^{\dag3}$\\
$^{1}$Beihang University, $^{2}$Tsinghua University, $^{3}$Nanjing University, $^{4}$Beijing Weilan Technology Co., Ltd.\\
}

\begin{document}
\maketitle

\renewcommand{\thefootnote}{\fnsymbol{footnote}}
\footnotetext[1]{Equal contribution.}
\footnotetext[2]{Corresponding author.}
\renewcommand{\thefootnote}{\arabic{footnote}}

\begin{abstract}
Foot contact is an important cue for human motion capture, understanding, and generation. Existing datasets tend to annotate dense foot contact using visual matching with thresholding or incorporating pressure signals. However, these approaches either suffer from low accuracy or are only designed for small-range and slow motion. There is still a lack of a vision-pressure multimodal dataset with large-range and fast human motion, as well as accurate and dense foot-contact annotation. 
To fill this gap, we propose a \textbf{M}ultimodal \textbf{M}oCap Dataset with \textbf{V}ision and \textbf{P}ressure sensors, named MMVP. MMVP provides accurate and dense plantar pressure signals synchronized with RGBD observations, which is especially useful for both plausible shape estimation, robust pose fitting without foot drifting, and accurate global translation tracking. 
To validate the dataset, we propose an \textit{RGBD-P SMPL fitting} method and also a monocular-video-based baseline framework, \textit{VP-MoCap}, for human motion capture. 
Experiments demonstrate that our RGBD-P SMPL Fitting results significantly outperform pure visual motion capture. Moreover, VP-MoCap outperforms SOTA methods in foot-contact and global translation estimation accuracy. 
We believe the configuration of the dataset and the baseline frameworks will stimulate the research in this direction and also provide a good reference for MoCap applications in various domains. 
Project page: \href{https://metaverse-ai-lab-thu.github.io/MMVP-Dataset/}{https://metaverse-ai-lab-thu.github.io/MMVP-Dataset/}
\end{abstract}    
\section{Introduction}
\label{sec:intro}
Human motion capture is an important foundation for motion analysis, behavior understanding, and pose generation, with a wide range of applications in AR/VR, disease diagnosis, robot manipulation, sports training, etc. 
In recent years, human motion capture based on computer vision technology has been developing rapidly, which has mainly benefited from the development of deep learning technology as well as, more importantly, the release of a large number of human motion capture datasets. 
Most of the existing datasets focus on human shape and pose annotations. 
Given pairs of image and shape and pose annotations for training, learning-based methods can infer plausible human motion from visual signals alone~\cite{kanazawa2018hmr,cai2024smpler,kolotouros2019spin,kocabas2020vibe,sun2021romp,zhang2021pymaf,pavlakos2018learning,kocabas2021pare,li2022cliff,guler2019holopose}. 
However, there are still many limitations when only using pose and shape annotations for supervision, resulting in artifacts such as foot drifting and erroneous global translation\&rotation, etc. 

\begin{figure}[t]
  \centering
   \includegraphics[width=0.98\linewidth]{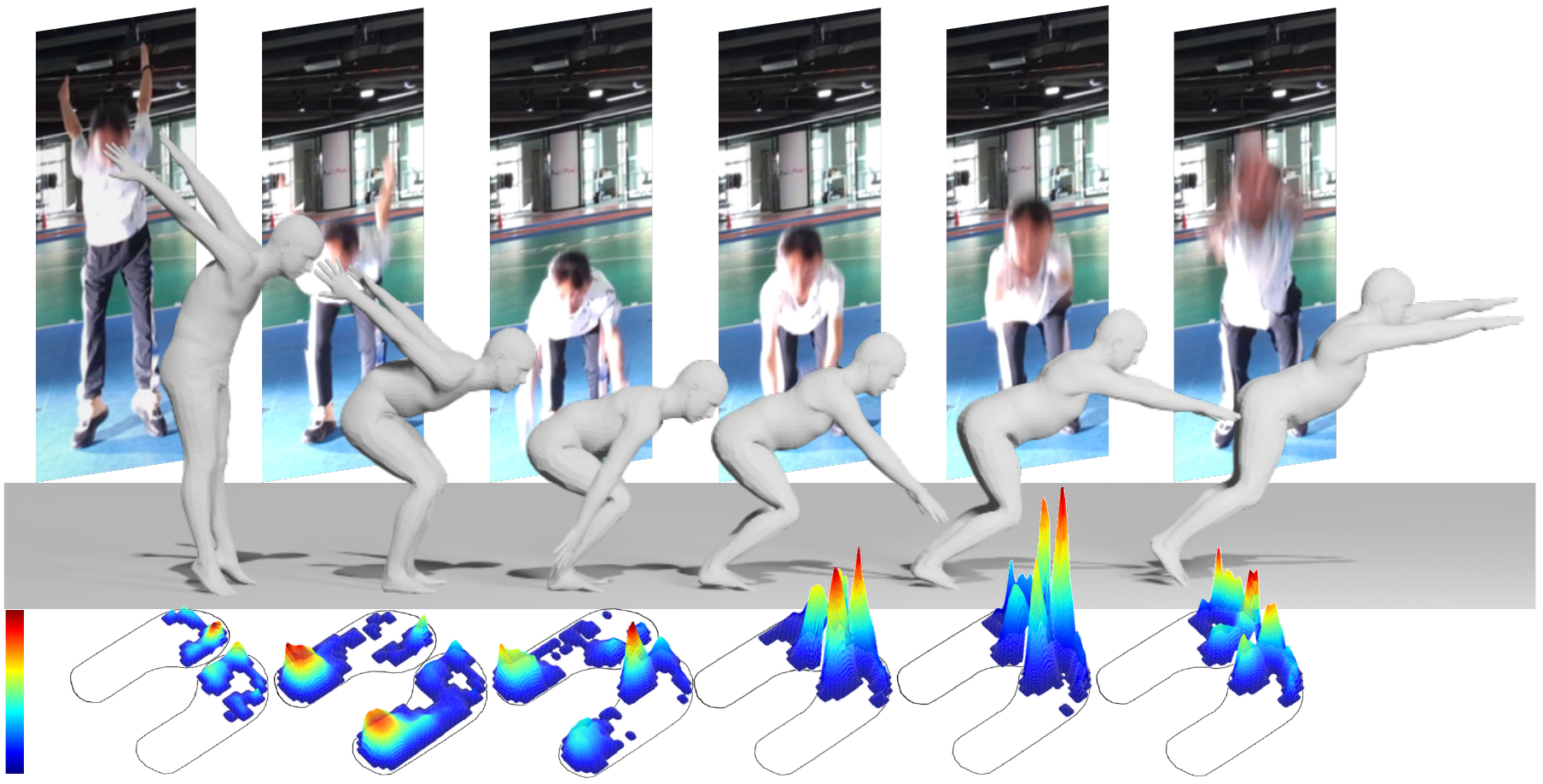}
   \caption{\textbf{MMVP} is a multimodal dataset that provides monocular RGBD video and accurate foot pressure (contact) of large-range and fast human motion.}
   \label{fig:main}
\end{figure}

To solve the above problem, recent datasets introduce additional annotations, mainly contacts, such as foot-contact, body-scene contact~\cite{hassan2021posa,huang2022rich} or self-contact~\cite{ fieraru2021self,Mueller2021tuch}, to enhance additional physical constraints for visual motion capture. Among these contact annotations, the most classical one is the foot-contact annotation. With foot contact annotation, Rempe \etal~\cite{rempe2020contact} has inferred the 0-1 foot-contact label for the foot joints from an image, which successfully restricts the foot drifting artifacts and realizes much more accurate global translation results. It is worth mentioning that the labeled contact information can not only improve the accuracy of motion capture but also be used as important cues to directly improve the rationality of motion analysis~\cite{shimada2020physcap,shimada2021neural,yi2022pip} and even motion generation~\cite{herzog2016structured}. 

However, it is very difficult to accurately annotate the contact. 
Most of the existing datasets use purely visual matching and distance thresholding for contact annotation ~\cite{hassan2021posa,hassan2019prox,huang2022rich}. 
Specifically, these datasets usually require a pre-scanning of the 3D scene, and then, based on the image observations, use an optimization method to fit the parameterized human model (SMPL~\cite{ loper2015smpl} or SMPL-X~\cite{pavlakos2019smplx}) to the 3D scene, and finally the GT (ground truth) contact was annotated by thresholding the distance between the vertices of the SMPL model and the 3D scene. 
Since this is an in-direct annotation strategy, the accuracy is inevitably affected by the pose estimation error, shape estimation error, and even scene scanning error. 
The situation becomes even worse under fast and large-range motion when the contact time is short and the foot motion is sophisticated. 
In short, when the purely visual fitting error is larger, it is impractical to set a fixed distance threshold for accurate and detailed foot contact annotation. 
As a result, the foot contact annotation of these datasets is not precise enough and the annotation granularity is relatively coarse. 

On the other end of the spectrum, in the field of health examination and clinical diagnosis, the use of specialized pressure sensors, such as the INSOLE, to obtain plantar pressure has been a proven and feasible option
~\cite{schreiber2019multimodal,van2022biomechanics,grouvel2023dataset}. 
However, these datasets often lack the simultaneous acquisition of visual signals, making it difficult to apply to CV tasks.
The most related multimodal datasets to ours are PSU-TMM100~\cite{scott2020image} and MOYO~\cite{tripathi2023IP}. 
Jesse \etal collected the PSU-TMM100 dataset with 2 RGB cameras and a pair of insole. However, this dataset is only designed for analyzing the stability of motion when humans perform Taiji Quan, so the movements are slow and the motion categories are relatively simple. 
Most recently, MOYO~\cite{tripathi2023IP} uses multiple RGB sensors with a fixed position force plate for capturing yoga and analyzing the stability. However, the force plate restricts the pressure acquisition area for capturing large-range motion. 

As a result, there is still a lack of a motion capture dataset that can simultaneously take into account a large range of motion, fast body movements, and visual-pressure synergistic acquisition, as well as accurate and dense foot-contact annotation. 
To fill this gap, we design a vision-pressure synchronized multimodal human motion capture system and acquire a new dataset named Multimodal MoCap dataset with Vision and Pressure Sensors (MMVP), as shown in \cref {fig:main}. 
Specifically, we captured single-view RGBD sequences and plantar pressure under a large range of rapid movements with sophisticated footwork. Based on the high-end insole sensor, we can obtain precise and dense foot pressure and contact annotations which are significantly more accurate and detailed than previous datasets. 
To fully utilize the synchronized RGBD and dense foot-contact signals for motion capture, we further design a novel RGBD-P SMPL fitting method based on multimodal input, which is far more accurate than vision-only methods, especially around foot regions. 
Finally, based on the new dataset and the fitted SMPLs, we propose and evaluate a baseline framework for monocular video-based human motion capture, VP-MoCap, which incorporates a per-vertex-level foot-pressure predictor (FPP-Net) and contact-based pose optimization strategy together to significantly improve the global translation and pose estimation accuracy. 

In summary, our contributions are: 
\begin{itemize}
\item We present MMVP, a novel multimodal human motion capture dataset that provides precise foot pressure and the most accurate dense contact annotations for large-range and rapid movements with detailed footwork. 

\item An RGBD-P SMPL fitting method is introduced to fully utilize the multimodal signals and achieve high-quality human motion capture results with plausible shape, stable foot movement, and accurate global translations. We also introduce a monocular RGB-based human motion capture baseline framework called VP-MoCap, which combines an FPP-Net and a joint optimization strategy.

\item Comparisons with SOTA SMPL fitting and monocular MoCap methods demonstrate the effectiveness of our dataset, RGBD-P fitting strategy, and the VP-MoCap framework. 

\end{itemize}
\section{Related Work}

\begin{table*}[htp]
  \centering
  \begin{tabular}{lcccc}
    \toprule
    Datasets & Visual Signal &Additional Signal&Additional Annotation\\
    \midrule
    PROX~\cite{hassan2019prox} & S.V. RGBD &N/A&Scene-contact&\\
    RICH~\cite{huang2022rich} & M.V. RGB  &N/A&Scene-contact&\\
    HumanSC3D~\cite{fieraru2021self} & M.V. RGB  &N/A&Self-contact&\\
    FlickrSC3D~\cite{fieraru2021self} & S.V. RGB &N/A&Self-contact&\\
    TUCH~\cite{Mueller2021tuch} & S.V. RGB &N/A&Self-contact&\\
    HOT~\cite{chen2023hot}& S.V. RGB &N/A&Object contact&\\
    TotalCapture~\cite{trumble2017total} & M.V. RGB &IMU&N/A&\\
    HPS~\cite{guzov2021human} & M.V. RGB &IMU&N/A&\\
    LiDARHuman26M~\cite{li2022lidarcap} & S.V. RGB &LiDAR&N/A&\\
    Schreiber \etal~\cite{schreiber2019multimodal}& N/A &EMG, force plate&Foot pressure\\
    Grouvel \etal~\cite{grouvel2023dataset} & N/A &IMU, insole, and force plate&Foot pressure\\
    Zee \etal~\cite{van2022biomechanics} & N/A &Insole&Foot pressure\\
    MoYo~\cite{tripathi2023IP} & M.V. RGBD &Force plate&Pressure and body contact\\
    \toprule
    Ours & S.V. RGBD &Insole&Foot pressure and contact\\
    \bottomrule
  \end{tabular}
  \caption{Comparison of existing multimodal human motion capture datasets. S.V.: single-view, M.V.: multi-view. Note that MoYo provides pressure and contact captured by a fixed-position force plate, so the capture range is limited. MMVP is the first dataset that provides accurate foot pressure and contact annotations with a large range and rapid body movements. }
  \label{tab:multimodal}
\end{table*}

\subsection{Multimodal MoCap Datasets}
The Human datasets~\cite{Ionescu2014h36m,Bogo2014FAUST,Alldieck2018snap,Mahmood2019amass,peng2021neural,yu2021function4d} have greatly promoted the development of pose detection. Nevertheless, these datasets only include visual signals and provide annotations for body shapes and poses. To further analyze human motion, multimodal datasets are needed. \cref{tab:multimodal} summarizes existing multimodal datasets.

Some datasets introduce additional interaction annotations, such as human-scene contact~\cite{hassan2021posa,huang2022rich}, self-contact~\cite{ fieraru2021self,Mueller2021tuch} or object contact~\cite{chen2023hot}, providing new insights for the study and analysis of human motion. These contacts can significantly reduce pose ambiguities in motion capture and enhance the understanding of visual images.
Whether manually annotated or based on thresholds, these labels are derived solely from visual information, making it difficult to guarantee accuracy.

Some datasets incorporate signals from other modalities. LiDAR is typically applied in large scenarios, and its signal modality is similar to vision~\cite{li2022lidarcap,dai2023sloper4d,yan2023cimi4d,yan2023cimi4d}. Inertial Measurement Units (IMUs) can measure the accelerations and orientations of body parts. Thus the combination of vision and IMUs can solve heavy occlusion or extreme illumination~\cite{trumble2017total,guzov2021human}. 
Pressure is a commonly used signal in human gait analysis, clinical rehabilitation assessment, and other biomedical fields. But medical datasets~\cite{grouvel2023dataset,schreiber2019multimodal,van2022biomechanics} often lack visual information.
TMM100~\cite{scott2020image} and MOYO~\cite{tripathi2023IP} are the most representative vision \& pressure datasets.
The pressure can provide accurate contact information and physical features.
However, both of these tend to favor slow and steady movements.
Currently, there is still a lack of datasets that synchronize pressure and visual data to describe a wide range of fast body movements.

\subsection{Human Pose Estimation}
Existing methods have made remarkable progress in human pose and shape estimation~\cite{joo2018total,zhang2021lightweight,liu2013markerless,li2011markerless,joo2015panoptic,guler2019holopose}.
While single image and monocular pose estimation has received more widespread attention~\cite{kanazawa2018hmr, bogo2016keepsmpl,pavlakos2019smplx,kocabas2020vibe,kolotouros2019spin,zhang2020object,li2021hybrik,feng2021pixie,li2022cliff}. 
However, these methods solely focus on the correlation between image features and human pose, neglecting the connection between subjects and their environment. Consequently, this limitation often results in visually and physically implausible outcomes in the world frame, such as jitter, penetration, sliding, and so on.

LIDAR is employed to localize the global position of distant targets. LiDARCap~\cite{li2022lidarcap}, HSC4D~\cite{dai2022hsc4d}, and Sloper4d~\cite{dai2023sloper4d} employ LIDAR data to estimate global human poses. CIMI4D~\cite{yan2023cimi4d} leverages LiDAR to acquire the human-scene contact information and regress the human pose for off-grounded motions. Due to the distance of the subjects, these methods often lack fine-grained motion details.

Some approaches have explored the fusion of RGB data with IMU data to enhance the accuracy and robustness of pose estimation. Matthew et al.~\cite{trumble2017total} incorporates Long Short-Term Memory (LSTM) networks to fuse IMU tracking data and visual pose embeddings to improve the accuracy of pose estimation. HybridFusion~\cite{zheng2018hybrfu} utilizes IMU data to address the occlusion problems in a real-time monocular MoCap system. Pan et al.~\cite{Pan2023Fusing} fuses RGB and IMU data using a complementary filter algorithm within the mocap framework to address the challenge of reliability in vision information.
EgoLocate~\cite{Ego2023Locate} leverages MoCap priors from inertial inputs to enhance the accuracy and robustness of the mocap system.

Contact provides strong interaction priors for pose estimation. Thanks to the contact-based datasets, many methods~\cite{hassan2021posa,huang2022rich,fieraru2021self} can estimate contact from a single image. 
The contact helps not only reduce ambiguity and achieve higher accuracy in pose estimation~\cite{Mueller2021tuch,hassan2019prox,fieraru2021self} but also provides dynamic constraints for rigid body dynamics-related methods~\cite{shimada2020physcap,shimada2021neural,yi2022pip}. 
In addition to determining contact, pressure can also provide force information. Zhang \etal~\cite{zhang2014leveraging} estimate human pose from three depth cameras and a pair of pressure-sensing shoes. Tripathi \etal~\cite{tripathi2023IP} enhance the stability in pose estimation by encouraging plausible floor contact
and overlapping the center of pressure and the
SMPL’s center of mass.
\section{MMVP Dataset}
\label{sec:vpd}

\subsection{Data Collection and Pre-processing}\label{sec:data_collection}

We collect a multimodal dataset containing synchronized RGBD video and foot pressure data. 
For visual data, we record the RGBD video using an Azure Kinect camera at a frame rate of 30Hz. 
For pressure data, we use Xsensor pressure insoles (HX 210-510). Each Xsensor pressure insole contains 242 independent pressure sensors that accurately capture the changes in pressure, with a frame rate of up to 150Hz. 
Due to the hardware limitation, we cannot synchronize the two streams of signals automatically, so we do manual synchronization before usage. 
The MMVP dataset covers up to 10 types of large-range and fast movements, including running, standing-long-jumping, and skipping, among other common types of exercises and dances. The majority of the subjects are teenagers, covering different age groups and both genders. For more details, please refer to the supplementary material.

\subsection{Calculate Dense Foot Contact}\label{sec:cal_foot_contact}

\begin{figure}[t]
  \centering
   \includegraphics[width=0.80\linewidth]{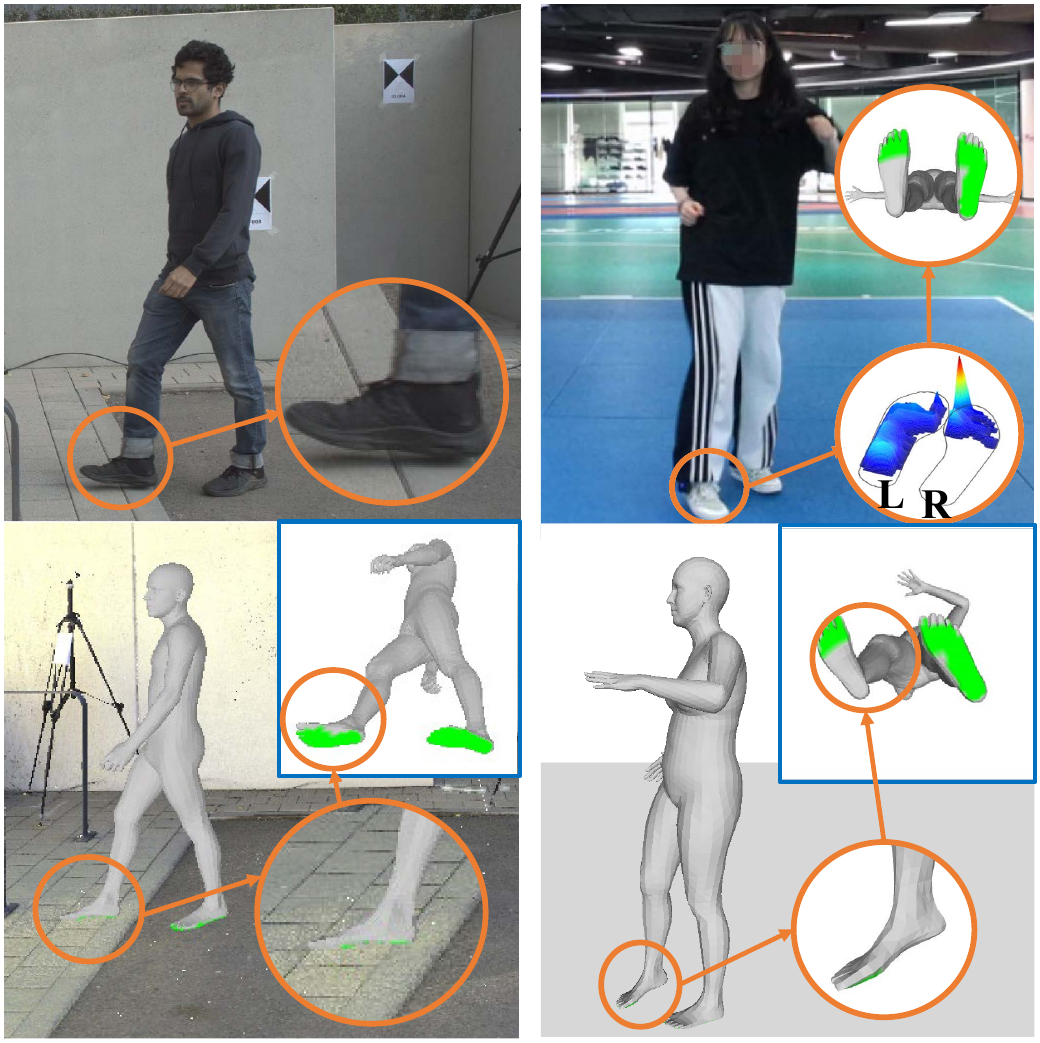}
   \caption{Comparison of the foot contact and pose annotations in RICH~\cite{huang2022rich} (left) and MMVP (right). RICH fits the SMPL model for annotating the foot contact by distance thresholding, while MMVP incorporates dense pressure and contact directly for much more accurate SMPL fitting results. Note that due to the vision-only SMPL fitting error of RICH, the right foot, which is hanging in the air, was annotated as full contact with the ground. }
   \label{fig:annotation}
\end{figure}

Unlike defining contact on joints~\cite{rempe2020contact}, dense contact~\cite{hassan2021posa} defines contact on the vertices of the human body model to describe detailed body-scene interaction. 
Existing methods~\cite{hassan2019prox,huang2022rich} usually first require scanning a 3D model of the scene and tracking the movements of subjects based on visual observations. Then annotate the dense contact by thresholding the distance between the body mesh and the scene. However, the inevitable pose estimation errors and scene reconstruction errors result in erroneous foot contact annotations, as shown in \cref{fig:annotation} (left). 

Benefiting from the highly accurate pressure insole, we can calculate precise and dense foot contact annotations. 
To ease the usage of the dense pressure and contact information, we define it on the surface of the SMPL~\cite{loper2015smpl}. We select 192 vertices in total on the feet of the SMPL model and the dense contact label can be defined as $C\in\mathbb{R}^{192}$. 

The foot pressure can vary dramatically depending on body weight and motion status. So it's hard to design a fixed-value threshold to annotate the dense contact.
Luckily, with insoles, it's easy for us to get the accurate body weight of different subjects by simply summarizing the pressure at the beginning when the subject is standing still. 
Therefore we normalized and mapped the pressure to $[0,1]$ as
\begin{equation}
  P_{\mathit{norm}} = \mathit{Sigmoid}(P/w_s),
  \label{eq:press_norm}
\end{equation}
where $P$, $P_{\mathit{norm}}$, $w_s$ are the originally captured pressure, normalized pressure, and body weight, respectively. 

\begin{figure}[t]
  \centering
   \includegraphics[width=\linewidth]{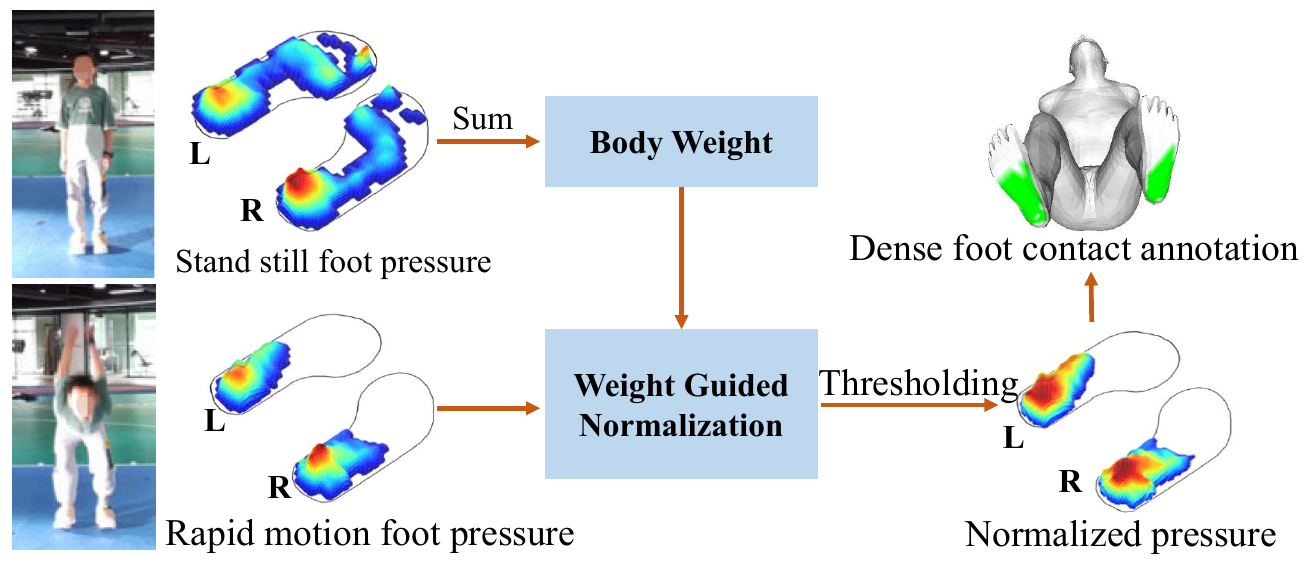}
   \caption{Illustration of the dense foot contact annotating method. From left to right are the reference image, original pressure, normalized pressure, and dense contact.}
   \label{fig:calc_cont}
\end{figure}

Finally, we empirically set the threshold to 0.5 to ensure a relatively strict contact annotation: 
\begin{eqnarray}
C= \begin{cases}
1,& \mbox{$P_{\mathit{norm}} \geq$ 0.5,}\\
0,& \mbox{$P_{\mathit{norm}} < $ 0.5.}\\
\end{cases}
\end{eqnarray}
\cref{fig:calc_cont} illustrates the calculation process of the dense foot contact annotation. 

\begin{figure*}
  \centering
   \includegraphics[width=0.72\linewidth]{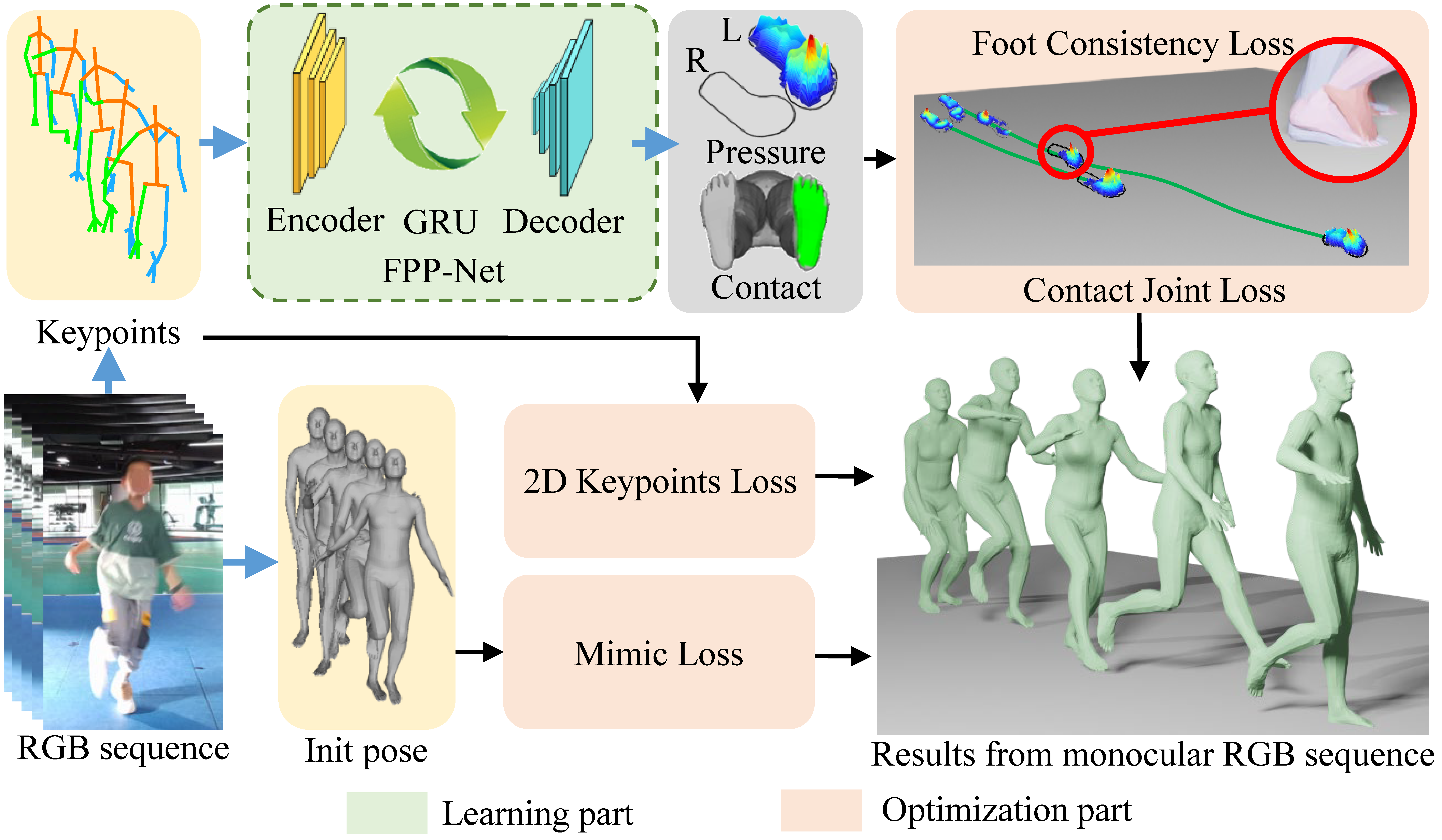}
   \caption{VP-MoCap pipeline. Given an RGB sequence, RTMPose~\cite{jiang2023rtmpose} and CLIFF~\cite{li2022cliff} are applied to detect 2D keypoints and regress the initial pose. FPP-Net predicts foot pressure distribution and dense foot contact with keypoint sequence. Guided by 
   foot contact, joint optimization is applied to estimate pose and trajectory. (Green represents the learning part, while orange represents the optimization part.)}
   \label{fig:pipeline}
\end{figure*}

\subsection{RGBD-P SMPL Fitting}
As illustrated in \cref{fig:annotation}, existing datasets calculate contact after fitting the body model to the pre-scanned scene with purely vision signals. Therefore, it is difficult to accurately maintain the interaction relationship between the human body and the scene during fitting. 
By constructing loss terms according to the ready-to-use contact information derived from the pressure signal, we can obtain more accurate pose and shape-fitting results.
Specifically, we use SMPL~\cite{loper2015smpl} to numerically represent the body shape and pose. The SMPL model takes pose parameters $\boldsymbol{\theta} \in \mathbb{R}^{72}$, shape parameters $\boldsymbol{\beta} \in \mathbb{R}^{10}$, global rotation $\boldsymbol{R} \in SO(3)$, and global translation $\boldsymbol{T} \in \mathbb{R}^{3}$ as input, producing a triangulated mesh of 6890 vertices. 
The $k$ joints of the model are represented as $\boldsymbol{J}(\boldsymbol{\theta}, \boldsymbol{\beta},\boldsymbol{R},\boldsymbol{T}) \in \mathbb{R}^{k \times 3}$. 
To guarantee the shape-fitting performance for teenagers with various body scales in our dataset, we follow the operation in AGORA dataset~\cite{patel2021agora}, which introduces a blending parameter $\alpha$ to balance the shape difference between children and adults.
According to AGORA, SMPL template $\boldsymbol{M}_{A}$ and SMIL~\cite{hesse2018learning} template $\boldsymbol{M}_{C}$ are blended as $M_{blend}=\alpha \boldsymbol{M}_{A} + (1 - \alpha) \boldsymbol{M}_{C}$. The vertices of the blended model are denoted as $\boldsymbol{V}$.

We optimize \cref{eq:tracking_loss} to fit the human model to the RGBD-P sequence with 3D depth fitting error $E_{depth}$ introduced in Doublefusion~\cite{yu2018doublefusion} to align the depth point cloud and SMPL surface vertices, 2D projection fitting error $E_{2d}$ and Gaussian Mixture Model(GMM) constraint $E_{GMM}$ following SMPLify~\cite{bogo2016keepsmpl}. 
\begin{equation}
\label{eq:tracking_loss}
    \begin{aligned}
    E(\boldsymbol{\theta},\boldsymbol{T}, \alpha, \boldsymbol{\beta})=&\lambda_{depth}E_{depth}+\lambda_{C\_dense}E_{C\_dense}+\\
    &\lambda_{2d}E_{2d}+\lambda_{C\_temp}E_{C\_temp}+\\
    &\lambda_{GMM}E_{GMM}. 
    \end{aligned}
\end{equation}

To fully utilize the annotated dense foot contact to eliminate foot-ground drifting, we introduce a dense contact loss $E_{C\_dense}$ in \cref{eq:contact_loss}. 
\begin{equation}
\label{eq:contact_loss}
    \begin{aligned}
        E_{C\_dense}=\sum_{C=1} \| \Pi_{floor}({\boldsymbol{V}_{foot}}) \|, 
    \end{aligned}
\end{equation}
where $\Pi_{floor}({\boldsymbol{V}_{foot}})$ denotes the distance between the ground and the foot surface vertex in contact.

Furthermore, to maintain the temporal smoothness of the foot fitting results during contact, we extend the human mesh to incorporate foot planes parallel to the pseudo ground. Each item in the plane corresponds to one vertex in the foot surface. These planes can be controlled in the same manner as human template vertices. Consequently, we design $E_{C\_temp}$ as \cref{eq:contact_temp_loss} to enhance the temporal consistency of the contacted plane vertices. Please refer to the supplementary video for more details. 

\begin{equation}
\label{eq:contact_temp_loss}
    \begin{aligned}
        E_{C\_temp}=\sum_{C_{t-1}=1\cap C_{t}=1} \left\|(\boldsymbol{V}_{t-1}-\boldsymbol{V}_{t})_{planes}\right\|_2, 
    \end{aligned}
\end{equation}
where $C_{t-1}=1\cap C_{t}=1$ denotes the intersection between the sets of foot plane vertices in frame $t-1$ and frame $t$ with a contact label of 1.
\section{Method: VP-MoCap}

\label{sec:method}
Previous monocular RGB-based human mesh recovery methods~\cite{li2022cliff} usually suffer from translation drifting, foot sliding, and foot-plane penetration, which make them unstable when handling relatively large-range and rapid movements. 
In the RGBD-P SMPL fitting section, we observed a much more accurate and plausible pose and translation by using depth and pressure signals for joint optimization. 
Thus, we further explore a more general question: given only monocular RGB sequences, can we also achieve stable and physically plausible human motion capture results for large-range and rapid movements with inferred dense contact and 3D scene?

Therefore, we first construct FPP-Net to estimate foot contact and then use Zoedepth~\cite{bhat2023zoedepth} to estimate the 3D ground from a monocular RGB sequence. Finally, with foot contact and ground depth, we try to obtain better pose and translation through optimization using only the monocular RGB sequence. The overall framework is summarized in \cref{fig:pipeline}.

\subsection{FPP-Net}

To estimate foot pressure distribution and foot contact from an RGB sequence, we design a foot pressure predictor (FPP-Net), which is illustrated in the green dashed box in \cref{fig:pipeline}. We assume that foot pressure and contact are more related to body pose and motion dynamic state. 
Thus we follow Physcap~\cite{shimada2020physcap} and Jesse \etal~\cite{scott2020image} to use sequential 2D keypoints for contact prediction. 
The goal of this strategy is to also decouple the network training process from the dataset-capturing environments (when compared with direct image regression methods like BSTRO~\cite{huang2022rich} and DECO~\cite{Tripathi2023deco}) thus significantly enhancing the generalization capacity.
The 2D keypoints sequence $\mathcal{J}^{2d}=\{j^{2d}_1, j^{2d}_2,...j^{2d}_t\}$ is detected by RTMPose~\cite{jiang2023rtmpose}.
The task can be formulated as
\begin{equation}
  \widetilde{C} = f(\mathcal{J}^{2d}).
  \label{eq:net_form}
\end{equation}

We first use a encoder $f_1$ to extract motion feature $\mathit{feat}^{motion}_t\in\mathbb{R}^{2048}$ for each frame $f_1(j^{2d}_1)$, $f_1(j^{2d}_2)$, ..., $f_1(j^{2d}_t)$. Then the motion feature is fed to a Gated Recurrent Unit (GRU) layer $f_2$ to yield temporal feature $\mathit{feat}^{temp}_t\in\mathbb{R}^{484}$. At last, a Multilayer Perceptron (MLP) is applied as the decoder to regress the final contact label $\widetilde{C}\in\mathbb{R}^{192}$.

Our training loss is 
\begin{equation}
 \mathcal{L} = \mathcal{L}_{cont}+\mathcal{L}_{press}.
  \label{eq:net_loss}
\end{equation}
Contact loss $\mathcal{L}_{cont}$ is the binary cross entropy loss between the ground truth contact and the predicted contact $\widetilde{C}$~\cite{huang2022rich}.
Pressure loss $\mathcal{L}_{press}$ is the mean squared error loss between the ground truth pressure and the predicted pressure~\cite{scott2020image}.

\subsection{Pose and Translation Optimization}
Since we pay more attention to human body pose and global translation, we do not optimize the intrinsic parameter $K$, shape $\tilde{\boldsymbol{\beta}}$, and global rotation $\tilde{\boldsymbol{R}}$; instead use weak-perspective projection to estimate $K$ and adopt   $\tilde{\boldsymbol{\beta}}$, $\tilde{\boldsymbol{R}}$, and initial pose $\tilde{\boldsymbol{\theta}}$ from CLIFF~\cite{li2022cliff}.
It is important to note that an image sequence corresponds to a specific individual. Therefore, we fix an average of $\tilde{\boldsymbol{\beta}}$ for the same person across the entire image sequence. To simplify the notation, we use $\boldsymbol{J}(\boldsymbol{\theta}, \boldsymbol{T})$ to represent $\boldsymbol{J}(\boldsymbol{\theta}, \tilde{\boldsymbol{\beta}},\tilde{\boldsymbol{R}},\boldsymbol{T})$.

\noindent\textbf{2D keypoints loss. }Due to large-range and rapid movements, our image sequence exhibits increased blur and occlusions, impacting the accuracy of the inferred pose parameter priors, $\tilde{\boldsymbol{\theta}}$. To improve the accuracy of $\boldsymbol{\theta}$, we construct 2D keypoints loss, which serves to penalize the weighted distance between the reprojected SMPL joints and the corresponding estimated 2D keypoints $\mathcal{J}^{2d}$: 

\begin{equation}
E_{2 d}=\sum_{i} c_{i} \| \mathcal{J}^{2d}_{i}-\Pi_{K}\boldsymbol{J}(\boldsymbol{\theta}, \boldsymbol{T})_{i}\|_{2}^{2}, 
  \label{eq:L2d}
\end{equation}
where $\Pi_{K}$ is the projection process from 3D to 2D through the intrinsic parameter $K$. The weight $c$ is representative of the confidence associated with each estimated 2D keypoint.

\noindent\textbf{Mimic loss. }We also define a mimic loss to quantify the similarity of the pose and the initial pose,
\begin{equation}
E_{p}=\|\boldsymbol{\theta}-\tilde{\boldsymbol{\theta}}\|_{2}^{2}.
  \label{eq:L2p}
\end{equation}

\noindent\textbf{Contact joint loss. } To recover the positional relationship between the human and the ground, we utilize the off-the-shelf depth estimation model ZoeDepth \cite{bhat2023zoedepth} to obtain a depth map for each RGB image and use the corresponding point cloud $\boldsymbol{P}$ to construct the 3D contact trace on the ground. Therefore, the spatial translation constraint term can be defined by penalizing the 3D distance between the SMPL foot joint $\boldsymbol{J}(\boldsymbol{\theta}, \boldsymbol{T})_{foot}$ and the contact point cloud $\boldsymbol{P}(\mathcal{J}^{2d}_{foot})$:

\begin{equation}
E_{3 d}=\sum \| \boldsymbol{P}(\mathcal{J}^{2d}_{foot})-\boldsymbol{J}(\boldsymbol{\theta}, \boldsymbol{T})_{foot}\|_{2}^{2}.
  \label{eq:L3d}
\end{equation}

\noindent\textbf{Foot consistency loss. }Meanwhile, we assume that the foot joints' position of two consecutive frames is consistently close during contact. At the same time, the jitter of 2D keypoints 
$\mathcal{J}^{2d}_{foot}$ will cause the corresponding contact joint on the ground to shift significantly. To solve the above problems, we define a temporal constraint term as

\begin{equation}
E_{t}=\sum \| \boldsymbol{J}(\boldsymbol{\theta}^{\boldsymbol{t}}, \boldsymbol{T}^{\boldsymbol{t}})_{foot}-\boldsymbol{J}(\boldsymbol{\theta}^{\boldsymbol{t-1}}, \boldsymbol{T}^{\boldsymbol{t-1}})_{foot}\|_{2}^{2},
  \label{eq:Lt}
\end{equation}

Combining all the above constraints \cref{eq:L2d,eq:L2p,eq:L3d,eq:Lt}, 
the total objective function 
can be summarized as:

\begin{equation}
\arg\min_{\boldsymbol{\theta}, \boldsymbol{T}} E_{2 d}+\lambda_{p}E_{p} + \lambda_{3d} E_{3 d}+ \lambda_{t} E_{t},
  \label{eq:opt_full}
\end{equation}
where $\lambda_{p}$, $\lambda_{3d}$ and $\lambda_{t}$ are corresponding weights.

\section{Experiments}
\label{sec:exp}

\subsection{Comparison of Ground Truth Registration}
We compare our RGBD-P fitting method with two existing RGBD methods: PROX~\cite{hassan2019prox} and LEMO~\cite{zhang2021learning}. Both of these methods consider human-scene contact in pose fitting. We employ the 3D depth error $E_{3d}$ to quantify the alignment between human models and depth observations.
To evaluate the interaction between the model and the scene, we calculate the model contact by setting a unified threshold and compare the Mean Foot-Contact Error (MFCE), F1 score, and IOU with the ground truth contact. \cref{tab:eval_gt} demonstrates the effectiveness of the proposed method.

\begin{table}[!ht]
  \centering
  \begin{tabular}{@{}lccccc@{}}
    \toprule
    Method & $E_{3d}$$\downarrow$ & M.$\downarrow$ & F1$\uparrow$  & IOU$\uparrow$\\
    \midrule
    PROX~\cite{hassan2019prox} & \cellcolor{Second}{0.045}  & {9.348} &\cellcolor{Second}{0.475} & \cellcolor{Second}{0.439}\\
    LEMO~\cite{zhang2021learning}[stage 2] &\cellcolor{Third}{0.150} &\cellcolor{Second} {8.199} & \cellcolor{Third}{0.423} & \cellcolor{Third}{0.433}\\
    LEMO[stage 3] & {0.175} & \cellcolor{Third}{8.231} & {0.377} & {0.399}\\
    \midrule
    Ours & \cellcolor{First}{0.043} & \cellcolor{First}{7.968} &\cellcolor{First}{0.518} &\cellcolor{First}{0.507}\\
    \bottomrule
  \end{tabular}
  \caption{Evaluation of pose and shape registration on MMVP. M.: MFCE.}
  \label{tab:eval_gt}
\end{table}

\begin{table}[!ht]
  \centering
  \begin{tabular}{@{}lccccc@{}}
    \toprule
    Method & prec.$\uparrow$ & recall$\uparrow$ & F1$\uparrow$ & IOU$\uparrow$ \\
    \midrule
    BSTRO~\cite{huang2022rich} & \cellcolor{Third}{0.212} & \cellcolor{Second}{0.755}  & \cellcolor{Third}{0.318} &\cellcolor{Third}{0.205}\\
    BSTRO[FT] & \cellcolor{Second}{0.521} & 0.531 & \cellcolor{Second}{0.496}&\cellcolor{Second}{0.496}\\
    DECO~\cite{Tripathi2023deco} & 0.112 & \cellcolor{First}{0.853}  & 0.191 &0.112\\
    \midrule
    Ours &  \cellcolor{First}{0.529} & \cellcolor{Third}{0.586} & \cellcolor{First}{0.532} &  \cellcolor{First}{0.522} \\
    \bottomrule
  \end{tabular}
  \caption{Quantitative comparison of contact estimation on MMVP-test.}
  \label{tab:eval_cont}
\end{table}

\begin{figure}[!ht]
  \centering
  \includegraphics[width=0.95\linewidth]{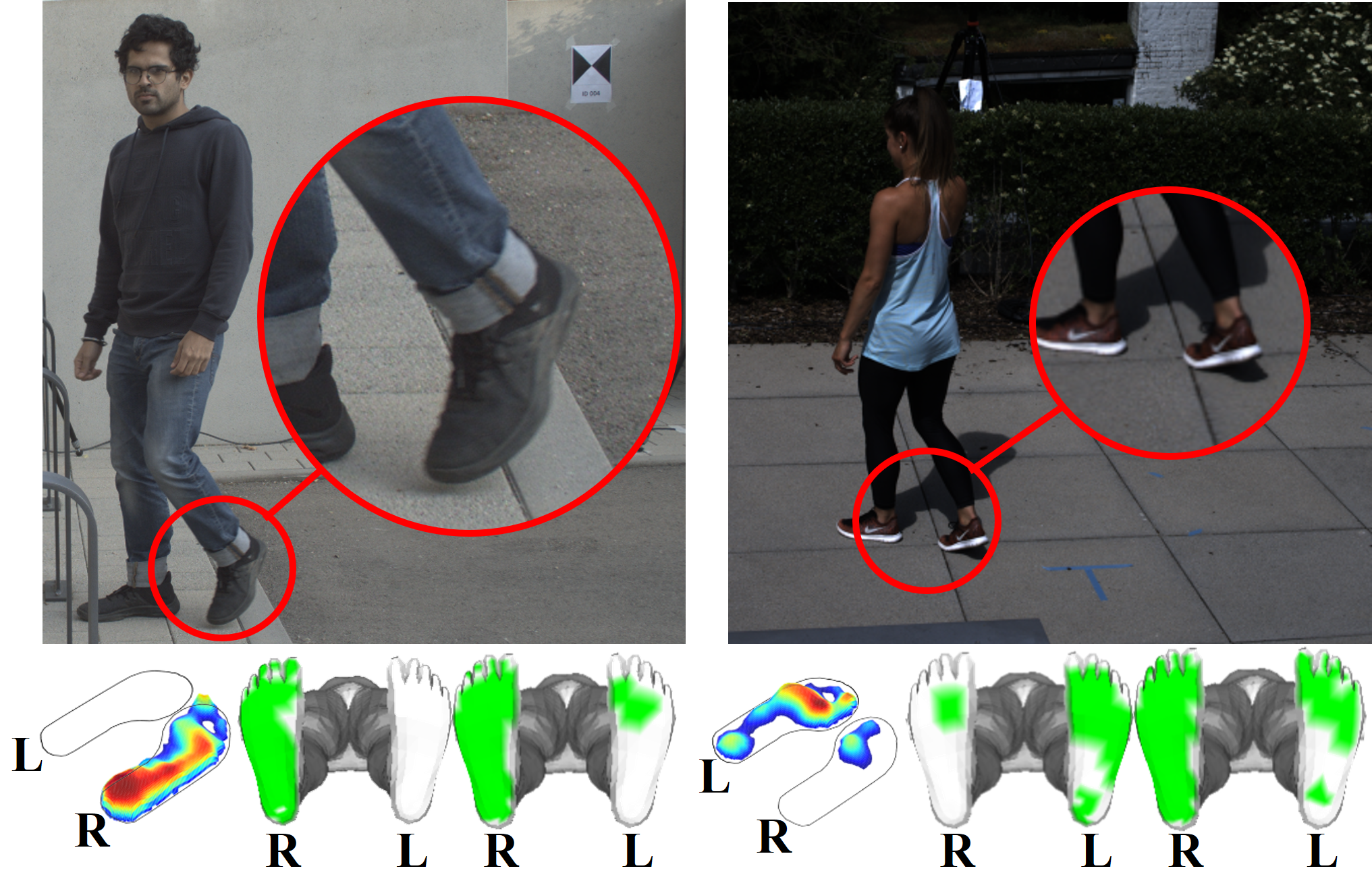}
   \caption{Qualitative comparison of contact estimation. From left to right, the second row includes: foot pressure distribution predicted by our method, foot contact predicted by our method, and contact predicted by BSTRO[FT].}
   \label{fig:eval_cont}
\end{figure}

\subsection{Comparison of Foot Contact Estimation}
We apply precision, recall, F1 score, and IOU as evaluation metrics and conduct a quantitative comparison with BSTRO~\cite{huang2022rich}, BSTRO[FT], and DECO~\cite{Tripathi2023deco} on MMVP-test (\cref{tab:eval_cont}). BSTRO[FT] represents the BSTRO fine-tuned on the MMVP training dataset. 
From \cref{tab:eval_cont}, it is evident that our method achieves the best performance. It is worth mentioning that BSTRO and DECO, due to contact annotations in their training dataset (illustrated in \cref{fig:annotation}), tend to predict feet as in contact for the majority of cases. As a result, their recall values are very high, while other metrics are undesirable. It is challenging to perceive the foot contact solely based on a single-frame image. 
By incorporating temporal motion information, we believe that's why the proposed method surpasses the fine-tuned BSTRO.

To validate the generalization, we also compare our method with BSTRO[FT] on public datasets. Due to the inaccurate contact labels in other datasets, we only conduct qualitative comparisons (\cref{fig:eval_cont}). It can be observed that the proposed method not only achieves more accurate contact predictions but also estimates reasonable foot pressure distributions. For more details, please refer to the supplementary material.
\begin{figure*}[t]
  \centering
   \includegraphics[width=0.97\linewidth]{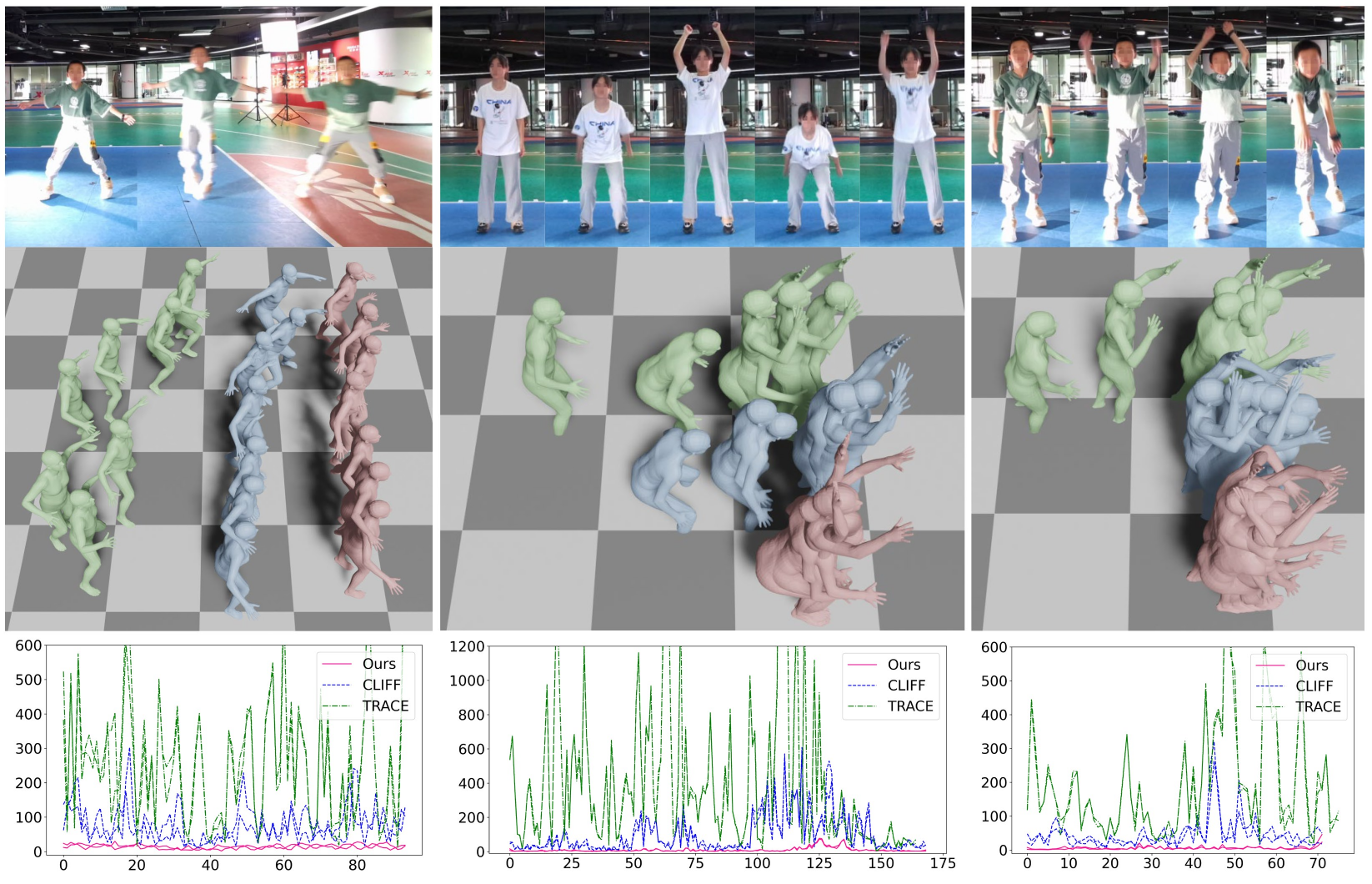}
   \caption{Qualitative comparison of 3D translation estimation in Side-Stepping (Left), Standing-Long-Jump (Middle) and Throwing-Medicine-Ball (Right) case. We estimate 3D translation stability and compare our results (Pink) with CLIFF~\cite{li2022cliff} (Blue) and TRACE~\cite{sun2023trace} (Green). The bottom row represents the acceleration of foot joints over time, which is used to measure the jitter of foot joints.}
   \label{fig:eval_trans}
\end{figure*}

\subsection{Pose and Translation Optimization Results}
\begin{table}
  \centering
  \begin{tabular}{@{}lccccc@{}}
    \toprule
    Methods & M. $\downarrow$ & PM.  $\downarrow$ & PVE $\downarrow$ & PF. $\downarrow$ & Traj $\downarrow$  \\
    \midrule
    CLIFF~\cite{li2022cliff} & \cellcolor{Third}{86.8} & 58.7  & \cellcolor{Third}{110.9} & 105.7 & 211.4 \\
    TRACE~\cite{sun2023trace} & 96.8 &  64.4 & 118.6 & 125.4 & 598.0\\
    SMPLer-X~\cite{cai2024smpler} & 92.2 & \cellcolor{First}{51.1}  & 112.9 & 121.8 & 679.7 \\
    \midrule
    Ours  & \cellcolor{First}{83.0} & \cellcolor{Second}56.0  & \cellcolor{Second}{110.6} & \cellcolor{First}{91.9} & \cellcolor{First}{129.3} \\
    (w/o $L_{t}$) & 87.0 & 57.9  & 114.0 & \cellcolor{Second}97.3 & \cellcolor{Second}133.4 \\
    (w/o $L_{3d}$, $L_{t} $) & \cellcolor{Second}{84.1} & \cellcolor{Third}{57.0}  & \cellcolor{First}{108.2} & \cellcolor{Third}{98.9} & \cellcolor{Third}{184.1} \\
    
    \bottomrule
  \end{tabular}
  \caption{Evaluation of pose and translation estimation on MMVP. M.: MPJPE, PM. : PMPJPE, PF. : PVE (Feet). }
  \label{tab:pose evaluation}
\end{table}
To assess the accuracy of our 3D pose estimation, we employ various metrics: MPJPE (Mean Per Joint Position Error), PMPJPE (Procrustes-aligned Mean Per Joint Position Error), PVE (Per Vertex Error), and PVE (Feet) which specifically focuses on the error of SMPL mesh vertices corresponding to the soles of the feet.
For evaluating translation estimation, we utilize the offset distance of the pelvis point relative to the starting position, denoted as Traj.

\noindent\textbf{Quantitative results. }
We conduct a comparative analysis of our method against CLIFF~\cite{li2022cliff}, TRACE~\cite{sun2023trace}, and SMPLer-X~\cite{cai2024smpler} on MMVP. 
As depicted in \cref{tab:pose evaluation}, our method exhibits satisfactory performance across all metrics. 
Notably, our method excels in Traj, indicating a significant improvement in acquiring more accurate and robust translation by leveraging foot contact for SMPL fitting.

\noindent\textbf{Qualitative results.} We compare 3D translation between our method, CLIFF~\cite{li2022cliff}, and TRACE~\cite{sun2023trace} on MMVP (\cref{fig:eval_trans}). In the Side-Stepping case, both CLIFF and TRACE exhibit drift in the depth direction, whereas our method demonstrates a consistent and stable translation in the depth direction. Similarly, in the Standing-Long-Jump and Throwing-Medicine-Ball case, the 3D translation estimated by CLIFF and TRACE manifests drift in the depth direction during the preparation stage. In contrast, our method avoids any depth-direction translation. At the bottom, we present the average absolute acceleration of foot joints which means the level of jitter. 
Our method achieves the smallest value, indicating superior foot stability.

\subsection{Ablation Studies} 
To investigate the impacts of $L_{3d}$ and $L_{t}$, we conduct an ablation experiment on the MMVP dataset.
As shown in \cref{tab:pose evaluation}, the omission of $L_{t}$ results in a marginal increase in all error metrics. This observation indicates that the temporal foot contact signals serve not only to improve the accuracy of translation but also to improve the accuracy of the pose. Without $L_{3d}$, MPJPE, PA-MPJPE, and PVE dropped slightly, but PVE (Feet) and Traj increased a lot. This is because when the 3D constraints of the feet and the ground are introduced, it will affect the 2D constraints to a certain extent and cause the accuracy of the pose to drop a little, but in exchange for the accuracy of the soles of the feet vertices and global translation. 
To attain enhanced translation, it is necessary to make a trade-off by sacrificing a certain degree of pose accuracy. 
Additionally, getting a better translation requires striking a balance between 2D and 3D constraints. 
\section{Conclusion}\label{sec:con}
The interaction between the human body and the scene has received increasing attention in recent years. However, most existing works struggle to provide precise contact annotations. To tackle this problem, we collect MMVP, a novel vision-pressure dataset with a wide range of rapid movements.
In addition, we present VP-MoCap, a monocular MoCap baseline framework that incorporates the prediction of dense pressure and contact for pose optimization.
Compared to solely relying on contact annotations, pressure signals contain richer dynamics information, particularly for fast and large-scale movements. We believe that our dataset holds great potential for future research.

\noindent\textbf{Limitations and future work:}
Our method is mainly designed for static viewpoints, so it remains challenging to handle moving camera configurations.
The existing MMVP has relatively homogeneous scenes, making it difficult for current visual networks to achieve scene generalization on this dataset.
In the future, we plan to collect data from various outdoor scenes to expand the dataset. Additionally, it would be an intriguing research endeavor to learn human body dynamics solely from pressure sequences.

\noindent\textbf{Acknowledgements:} This work was supported in part by the National Key R\&D Program of China under Grant No.2022YFF0902201, the NSFC No.62171255, Xtep Science Lab, the Tsinghua University-Joint research and development project under Grant R24119F0 JCLFT-Phase 1, the “Ligiht Field Generic Technology Platform" (Z23111000290000) of Beijing Municipal Science and Technology Commission, the Aeronautical Science Fund under Grant 20230023051001, the Guoqiang Institute of Tsinghua University under Grant No.2021GQG0001.
{
    \small
    \bibliographystyle{ieeenat_fullname}
    \bibliography{main}
}

\clearpage
\setcounter{page}{1}
\maketitlesupplementary

\section{Dataset Details} 
\begin{figure}[!htbp]
  \centering
   \includegraphics[width=\linewidth]{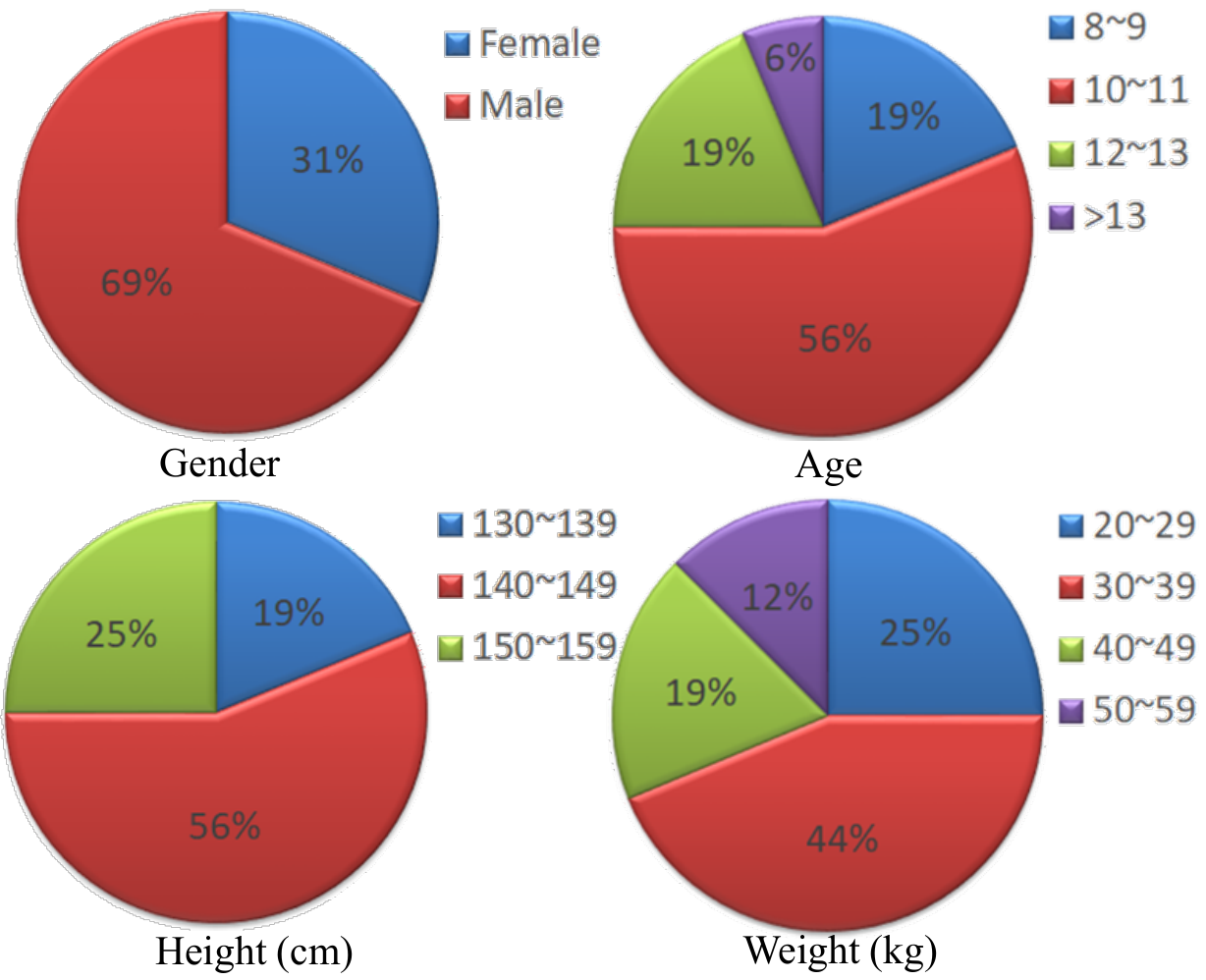}
   \caption{Statistics of MMVP subjects.}
   \label{fig:statistics}
\end{figure}
\noindent \textbf{Synchronization:} 
We uesed an Azure Kinect camera\footnote{https://azure.microsoft.com/en-us/products/kinect-dk/} and Xsensor pressure insoles (HX 210-510)\footnote{https://www.xsensor.com/solutions-and-platform/human-performance/gait-motion-insoles} to capture the data. 
Due to this pressure insole's lack of automatic synchronization support, we have designed a coarse-to-fine synchronization method. The pressing switch controls the bulb.
When the switch is pressed with the foot, the bulb illuminates, and the insole detects the pressure simultaneously, achieving coarse synchronization. Manual fine synchronization is performed by observing the sudden changes in pressure on the insole, such as landing after a long jump.

\noindent \textbf{Statistics:} The dataset consists of a total of 16 subjects, with 11 males and 5 females. 
Key statistics (gender, age, height, and weight) of the subjects are shown in \cref{fig:statistics}.
Synchronized RGBD frames, pressure data, and registrations are more than 44k frames.
The dataset consists of high-speed and large-range movements, including skipping rope, long jump, ball throwing, side stepping, running, and dancing.

\noindent \textbf{Ethics:} The subjects in MMVP are well-informed and have voluntarily signed legal agreements to allow the data to be made public for research purposes.

\section{RGBD-P fitting}
\begin{figure}[!htbp]
  \centering
   \includegraphics[width=1.0\linewidth]{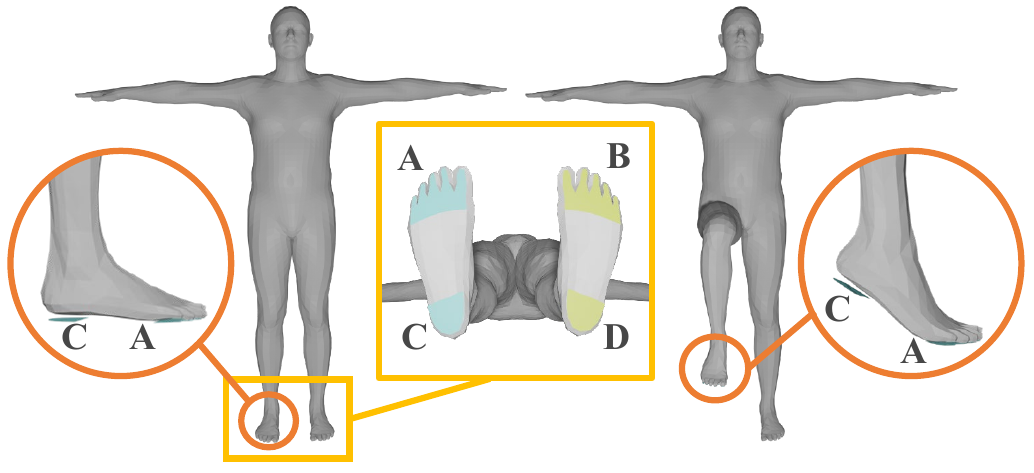}
   \caption{foot plane details.}
   \label{fig:foot_plane_details}
\end{figure}
\begin{figure}[!htbp]
  \centering
   \includegraphics[width=0.75\linewidth]{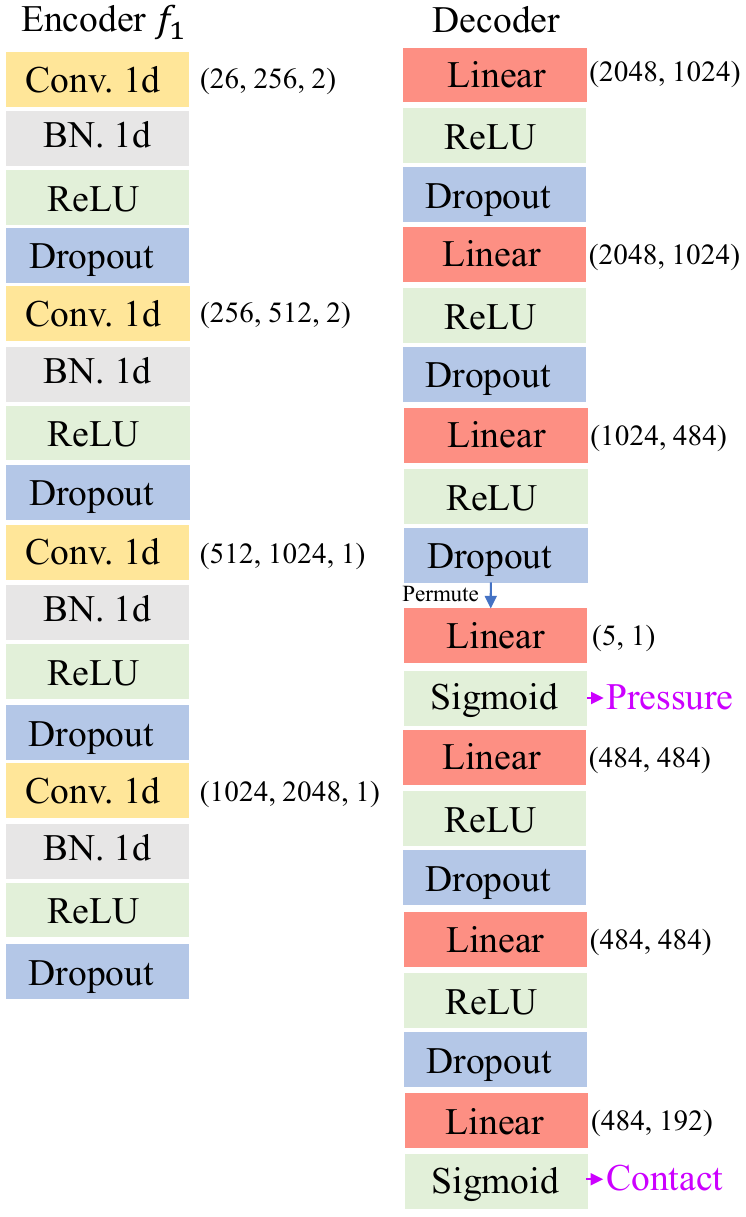}
   \caption{FPP-Net details.}
   \label{fig:net_details}
\end{figure}
\begin{figure*}[!ht]
  \centering
   \includegraphics[width=0.95\linewidth]{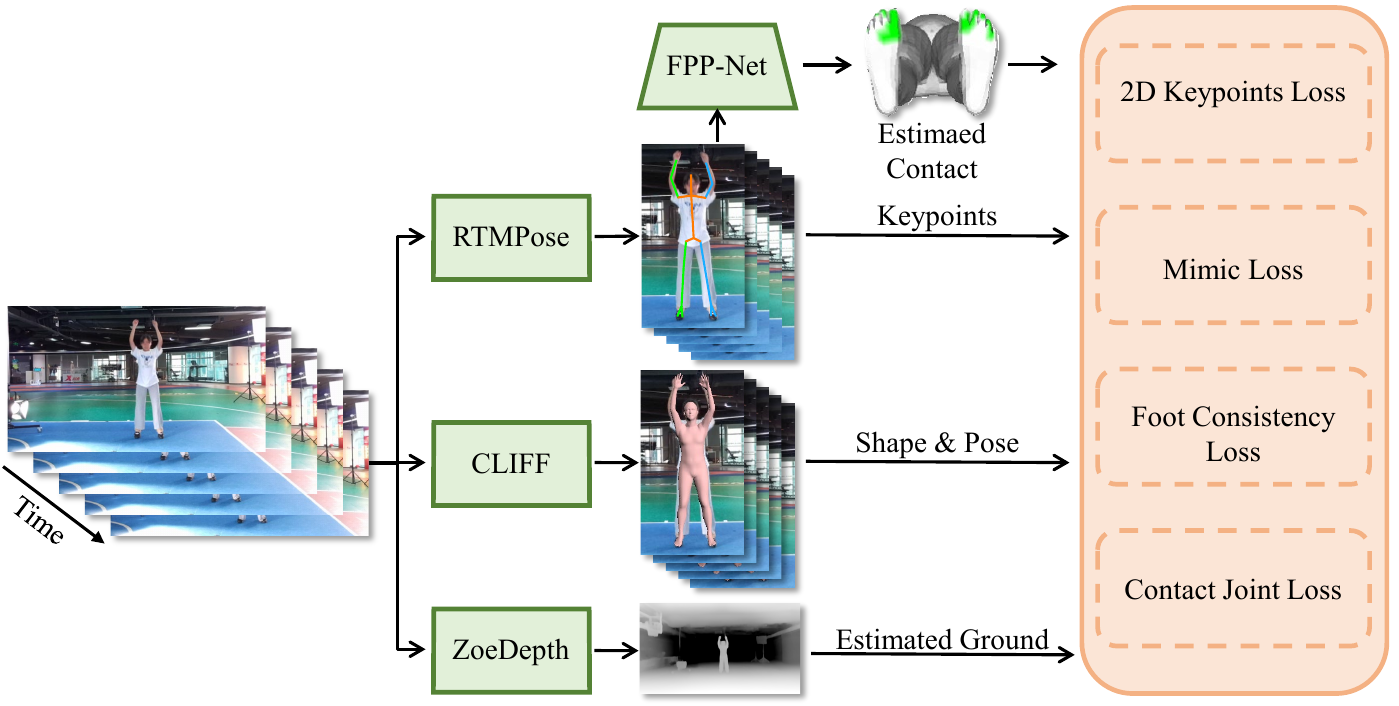}
   \caption{Flowchart of pose and translation optimization.}
   \label{fig:flowchart}
\end{figure*}
\subsection{Foot Contact Temporal Loss}
We design four-foot planes derived from SMPL template mesh, as shown in \cref{fig:foot_plane_details}.
By projecting the foot vertices of the template SMPL model onto the ground, we acquire the anterior and posterior foot planes. 
The points on foot planes, which are associated with SMPL foot vertices, can also be controlled by SMPL parameters.

When computing temporal contact constraints, we first convert the contact of SMPL vertices into contact of points on the plane.
Then, we select the points in each plane whose spatial position should remain static between adjacent frames. Subsequently, we calculate the L2 loss of the position of these selected points between frame $t-1$ and frame $t$. By minimizing this loss, we ensure that the selected points on the foot surface maintain consistent spatial positions over time.

When describing the state of the foot, foot planes can be more effective compared to foot surface vertices. These planes can provide valuable information about the orientation and inclination of the foot relative to the ground, which can aid in accurately controlling the foot pose. By utilizing this information from foot planes, we can enhance the precision and accuracy of foot pose control. 

\subsection{RGBD-P Fitting Implementation Details}
The optimization steps for ground truth generating can be divided into three parts: shape fitting, pose initialization, and pose tracking.

The first step is to fit the shape of the human body. We optimize the shape-related parameters $(\alpha, \beta)$ for each character. Specifically, we use the corresponding A-pose sequence from the dataset to perform this optimization.

The second step is to fit the initial pose parameter for each action sequence. For each action sequence, we select a specific frame, denoted as $t_0$, as the initial moment for pose tracking. Utilizing the shape-related parameters $(\alpha, \beta)$ calculated in the first step, we optimize the pose parameters $(\theta_{0}, \boldsymbol{T}_{0})$ for frame $t_0$.

The third step is to generate pose ground truth of each motion sequence. We leverage the shape-related parameters $(\alpha, \beta)$ and the initial pose parameters $(\theta_{0}, \boldsymbol{T}_{0})$ to estimate the pose for the entire motion sequence. During our tracking process at frame $t$, $(\theta_{t-1}, \boldsymbol{T}_{t-1})$ will be applied in fitting the pose parameters $(\theta_{t}, \boldsymbol{T}_{t})$.

The relevant methods are implemented using Python. We utilize Adam to optimize \cref{eq:opt_full} within PyTorch.

\subsection{RGBD-P fitting Comparison Details}
We compare our results with PROX~\footnote{https://github.com/mohamedhassanmus/prox} and LEMO~\footnote{https://github.com/sanweiliti/LEMO} on MMVP Dataset. All configs are set as their default values. We use four metrics to evaluate fitting performance: $E_{3d}$, Mean Foot-Contact Error (MFCE), F1 score, and IOU. For the last three metrics, given a fitting result, we set a fixed threshold to calculate the foot contact $C_{fitting}$ as POSA~\cite{hassan2021posa}. We calculate the F1 score and IOU between $C_{fitting}$ and the ground truth contact $C$. The Mean Foot-Contact Error (MFCE) is calculated as $\|C_{fitting} - C\|_2$.

Compared to PROX, our method considers body shape differences between adults and children, ensuring shape parameter consistency in one motion sequence. PROX sets a distance threshold between foot and ground, which may result in sudden motion. To avoid this, our method leverages dense foot-ground distance as loss function in optimization. Based on the fitting results of PROX, LEMO considers temporal motion changing and contact consistency, however, it neglects the alignment with depth data, which can lead to difficulties in accurately estimating global position. 


\section{VP-MoCap}
\subsection{FPP-Net}
\noindent\textbf{Network details.} We follow PhysCap~\footnote{https://github.com/soshishimada/PhysCap\_demo\_release} and Jesse \etal~\cite{scott2020image} to construct our network. The details are shown in \cref{fig:net_details}.
The numbers next to "Conv.1d" represent the input dimension, output dimension, and kernel size. The numbers next to "Linear" represent the input dimension and output dimension. The outputs of FPP-Net are marked in purple.
We train our network for 20 hours (900 epochs). The initial learning rate is $10^{-4}$. The optimizer is Adam.

\noindent\textbf{Dataset split.} We split 16 subjects in MMVP into 13 and 3 for training and testing, respectively.

\noindent\textbf{BSTRO fintune details.} We fintune BSTRO\footnote{https://github.com/paulchhuang/bstro} with MMVP-test on their pre-trained model. All configurations are set to default values.

\subsection{Pose and Translation Optimization Details}

As shown in \cref{fig:flowchart}, our pose and translation optimization is constructed using several inputs, including the estimated contact, human body keypoints, initial shape and pose, and estimated ground. 
The estimated contact is estimated through FPP-Net, utilizing human body keypoints as input, which are estimated by RTMPose~\cite{jiang2023rtmpose}. The initial shape and pose are estimated by the off-the-shelf method CLIFF~\cite{li2022cliff}. Additionally, the estimated ground is obtained through the application of ZoeDepth~\cite{bhat2023zoedepth}. We used VPoser~\cite{pavlakos2019expressive} as human body prior.

\subsection{ Pose and Translation Optimization Comparison Details}
We unify the focal length across our method and other comparative methods, maintaining the remaining settings of CLIFF~\footnote{https://github.com/huawei-noah/noah-research/tree/master/CLIFF}, TRACE~\footnote{https://github.com/Arthur151/ROMP/tree/master/simple\_romp/trace2} and SMPLer-X~\footnote{https://github.com/caizhongang/SMPLer-X} consistent with default parameters.

\begin{table}[!htbp]\footnotesize
  \centering
  \begin{tabular}{@{}clccccc@{}}
    \toprule
     & Methods & M. $\downarrow$ & PM.  $\downarrow$ & PVE $\downarrow$ & PF. $\downarrow$ & Traj $\downarrow$  \\
    \midrule
    \multirow{4}{*}{\rotatebox{90}{3DPW}} & CLIFF & \cellcolor{Second}{85.2} & \cellcolor{First}24.6  & \cellcolor{Second}{108.6} & \cellcolor{Second}{153.2} & - \\
    & TRACE & 99.4 & {37.2}  & 127.0 & 185.0 & -  \\
    & SMPLer-X & 109.3 &  \cellcolor{Second}30.1 & 137.2 & 197.1 & -\\
   &  Ours  & \cellcolor{First}{81.6} & 41.3  & \cellcolor{First}{106.4} & \cellcolor{First}{129.4} & - \\
    \midrule

    \multirow{4}{*}{\rotatebox{90}{EMDB}} & CLIFF & \cellcolor{Second}{93.8} & 39.7  & {129.0} & \cellcolor{Second}146.7 & \cellcolor{Second}105.2 \\
    & TRACE & 103.0 & {41.6}  & 133.4 & 197.8 & 114.5  \\
    & SMPLer-X & 95.8 &  \cellcolor{First}35.6 & \cellcolor{Second}127.0& 160.7 & 4943.0\\
   &  Ours  & \cellcolor{First}{89.0} & \cellcolor{Second}37.3  & \cellcolor{First}{125.9} & \cellcolor{First}{115.1} & \cellcolor{First}58.8 \\
    \bottomrule
  \end{tabular}
  \caption{Evaluation of pose and translation estimation on 3DPW and EMDB.M.: MPJPE, PM. : PMPJPE, PF. : PVE (Feet).}
  \label{tab:pose_evaluation}
\end{table}
\begin{figure}[!htbp]
  \centering
   \includegraphics[width=1.0\linewidth]{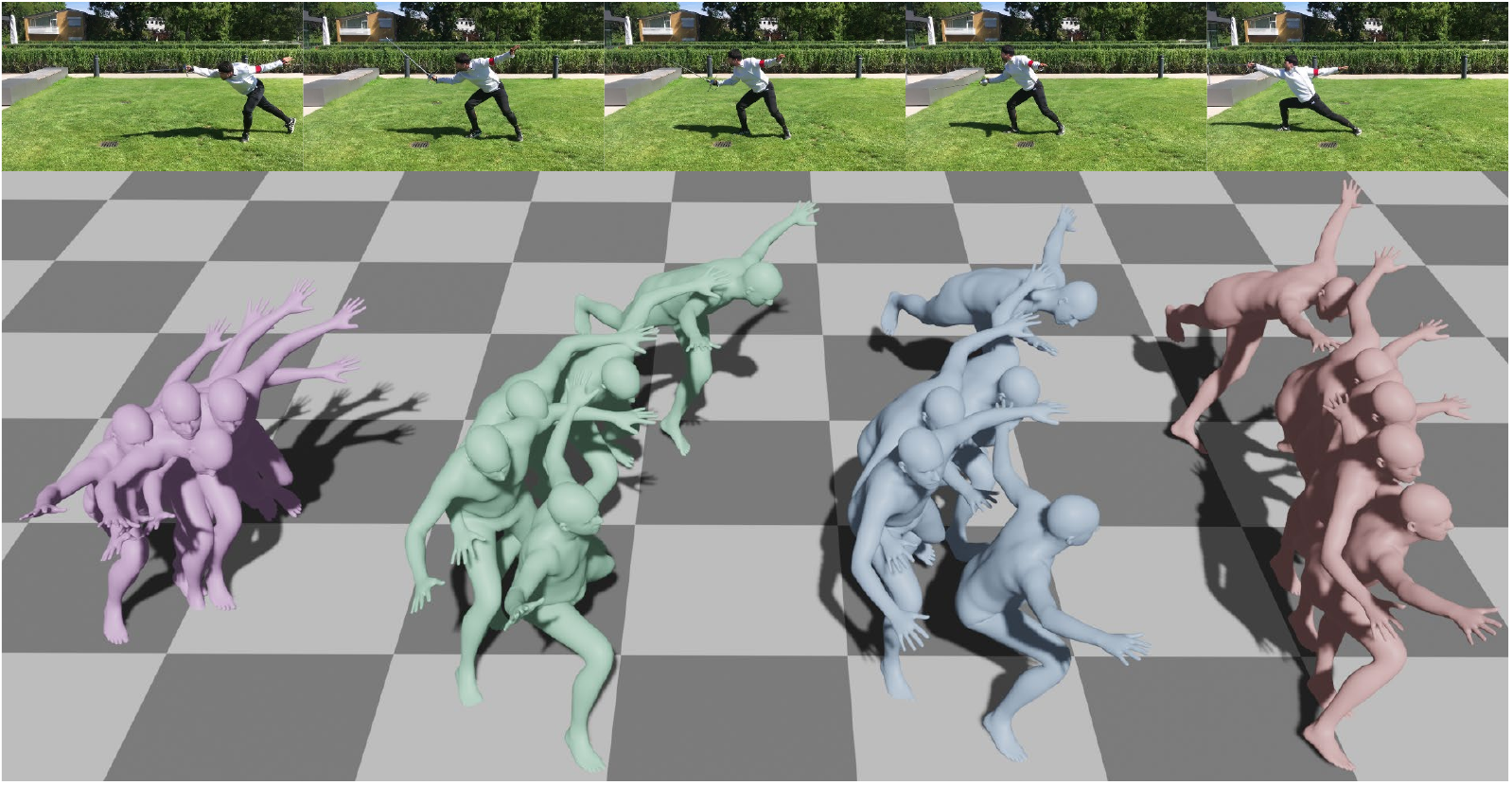}
   \caption{Comparison of 3D translation estimation in 3DPW. 
   From left to right: SMPLer-X (Purple), TRACE (Green), CLIFF (Blue), and our method (Pink).
   }
   \label{fig:dpw3_compare}
\end{figure}
\subsection{More Results on In-the-Wild Datasets}
We compared the results on in-the-wild datasets, 3DPW~\cite{von2018recovering} and EMDB~\cite{kaufmann2023emdb}. 
Our method is mainly designed for videos from static viewpoints, so it remains challenging to handle moving camera configurations. 
Therefore, we selected sequences with relatively static camera movements in 3DPW and EMDB (1K frames from 3DPW and 2.5K frames from EMDB) for conducting the comparison. 
Note that 3DPW does not provide GT global trajectory annotations for evaluation. 
As shown in \cref{tab:pose_evaluation} and \cref{fig:dpw3_compare}, our method achieved significant improvement in global translation estimation on EMDB (3DPW does not provide GT trans) and also produces slightly better pose estimation results.

\end{document}